\documentclass[letterpaper, 10pt, twocolumn]{article}
\usepackage{clean}
\shorttitle{MPC - HRI}
\newcommand{\appendices}[1]{#1} 

\usepackage{cite} 

\usepackage{graphicx}
\usepackage{caption}
\usepackage{subcaption}
\usepackage{amsmath,amssymb,amsfonts}
\usepackage{algorithm2e}
\usepackage{xcolor}
\usepackage[colorlinks=true, linkcolor=black, urlcolor=cyan, filecolor=black, citecolor=black]{hyperref}
\usepackage{url}
\usepackage{amsmath,bm}
\usepackage{multicol, multirow, makecell} % For table magic
\usepackage{comment}
\usepackage{algorithm2e}
\usepackage{xcolor}   
\usepackage{soul} % use \st{} for strikethrough
\newcommand{\rev}[1]{{\color{red} #1}} 

\renewcommand{\rev}[1]{{\color{black} #1}}
\renewcommand{\st}[1]{}

\begin{document}

\title{\LARGE \bf Planning Human-Robot Co-manipulation with Human Motor Control Objectives and Multi-component Reaching Strategies
}
\author{Kevin Haninger$^1$ and Luka Peternel$^2$
\thanks{This work was supported by the European Union's Horizon 2020 research and innovation programme under grant agreement No. 101058521 — CONVERGING.}% <-this % stops a space
\thanks{$^{1}$Department of Automation, Fraunhofer IPK, Berlin, Germany
        (e-mail:~kevin.haninger@ipk.fraunhofer.de)}
\thanks{$^{2}$Department of Cognitive Robotics, Delft Univerity of Technology, Delft, The Netherlands (e-mail:~l.peternel@tudelft.nl)}
\thanks{Code available at \url{https://gitlab.cc-asp.fraunhofer.de/hanikevi/hri-mpc}, video  at \url{https://youtu.be/9caL9bBedA0}}
}

\maketitle

\begin{abstract}
For successful goal-directed human-robot interaction, the robot should adapt to the intentions and actions of the collaborating human. This can be supported by \st{biomechanical}\rev{musculoskeletal} or data-driven \rev{human} models, where the former are limited to lower-level functioning such as ergonomics, and the latter have limited generalizability or data efficiency. What is missing, is the inclusion of human motor control models that can provide generalizable human behavior estimates and integrate into robot planning methods. We use well-studied models from human motor control based on the speed-accuracy and cost-benefit trade-offs to plan collaborative robot motions. In these models, the human trajectory minimizes an objective function, a formulation we adapt to numerical trajectory optimization. This can then be extended with constraints and new variables to realize collaborative motion planning and goal estimation. We deploy this model, as well as a multi-component movement strategy, in physical collaboration with uncertain goal-reaching and synchronized motion tasks, showing the ability of the approach to produce human-like trajectories over a range of conditions.  
\end{abstract}

\section{Introduction}
Human-robot collaboration is one of the key aspects of the seamless integration of robots into our daily lives. The principle of human-robot collaboration combines the human exceptional cognitive capabilities with robot strength and precision~\cite{ajoudani2018progress, fitts1951human}. Nevertheless, to do this effectively, it is important that the robot understands in real-time the states and intentions of the human partner, and acts in a legible way. The way to do so is the inclusion of human modeling into physical human-robot interaction control~\cite{fang2023human}.

In recent years, many studies have incorporated human musculoskeletal models into robot control systems to account for ergonomic factors, such as minimizing joint torques and muscle fatigue, with the goal of increasing work efficiency and preventing musculoskeletal injuries~\cite{lorenzini2023ergonomic}. While physical ergonomics is one of the important aspects of Human-robot collaboration, it is also crucial to be able to predict human behavior. The common communication channels for intention detection are based on vision~\cite{agravante2014collaborative}, haptic interaction~\cite{duchaine2012stable,donner2016cooperative,haninger2023model}, and biosignals, such as muscle activity~\cite{peternel2017human,bi2019review} or brain activity~\cite{lyu2022coordinating}. %sarac2013brain

The measurements from various channels then have to be interpreted by a model that can predict human behavior, so that the robot can generate appropriate collaborative actions. For haptic communication, the intention prediction model can be based on a simple impedance model that transforms the measured forces into predicted motion~\cite{duchaine2012stable,roveda2020model}. While this simplicity comes with good robustness, it is limited to encoding more primitive actions and behavior. On the other hand, more complex collaborative behavior can be encoded by machine learning methods, such as Bounded-Memory Adaptation Models~\cite{nikolaidis2017human}, Inverse Reinforcement Learning~\cite{losey2022physical}, Gaussian Processes (GP)~\cite{haninger2023model}, Dynamic Movement Primitives~\cite{peternel2018robot}, and ensemble
Bayesian Interaction Primitives~\cite{clark2022learning}. Nevertheless, these approaches rely on \st{learning processes and}\rev{human interaction} data which might not always be available, practical, or generalizable.

\begin{figure}[!t]
    \includegraphics[width=\columnwidth]{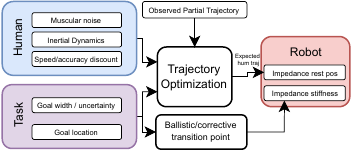}
    \caption{Proposed human motor control models (blue), combined with task parameters (purple) into a trajectory optimization problem to produce the robot trajectory.  This base model can be extended with observed motion or transitions between ballistic and corrective motion to build collaborative scenarios.}
    \label{fig:arch}
\end{figure}

An alternative to machine learning is to use computational models based on our understanding of human motor control. Human motor control studies developed various computational models of how the human central nervous system generates goal-directed movements, which are based on optimizing specific aspects. One important aspect the model can optimize for is the speed-accuracy trade-off~\cite{tanaka2006optimization,guigon2008computational}, which is described by the well-known Fitts' law~\cite{fitts1954information}. Since human neural signals are subject to noise, which is amplified by muscle activity~\cite{wolpert1998multiple}, faster movements that require more muscle activations tend to be less precise. Another important aspect of human motor control that the model can optimize is the cost-benefit trade-off~\cite{rigoux2012model,berret2016don,shadmehr2016representation}. The human central nervous system (CNS) wants to minimize the metabolic cost, which would require slower movements with less muscle activity. However, at the same time, spending more time on a movement is perceived as less beneficial~\cite{shadmehr2019movement}. A computational model can also combine the two trade-offs~\cite{peternel2017unifying}.

Furthermore, in more complex situations, the CNS can split movements into sub-movements to optimize them locally for different purposes~\cite{elliott2001century,peternel2019target}. When reaching for a more distant goal, the CNS tends to make the initial (ballistic) sub-movement faster and less precise to get into the vicinity of the goal as fast as possible, while the final (corrective) sub-movement is slower and more precise to ensure the goal is reached~\cite{elliott2001century,peternel2019target}. \rev{The proposed model considers that each sub-movement uses the speed-accuracy and cost-benefit trade-offs with different parameters depending on the purpose.} \st{This way speed-accuracy and cost-benefit trade-offs can be accounted for also on a higher level by splitting the optimization based on the purpose of each sub-movement in the overall scheme.}

Computational models of the CNS can directly integrate the understanding of human motor control into a collaborative robot control system for behavior prediction and optimization of the robot's actions. This can provide a principled way to generate trajectories that are compatible with or exploit natural human motion and reduce the need for machine learning data. \rev{ While autonomous robots do not need to operate according to human movement principles, in human-robot interaction, it is crucial for the robot to understand human motion and act according to human natural movement and limitations.}

A few existing studies in robotics incorporated some computational models into the robot control system using optimal control~\cite{jarrasse2012framework,liu2023follow}, goal inference~\cite{takagi2017physically,takagi2020flexible} or minimum-jerk principles~\cite{dallard2023synchronized}. However, these are not based on the standard or the latest accepted models from human motor control such as speed-accuracy and cost-benefit trade-offs, which \rev{are more accurate in neuroscience studies} \st{have been rigorously studied in neuroscience}~\cite{tanaka2006optimization,guigon2008computational,rigoux2012model,berret2016don,shadmehr2016representation,shadmehr2019movement,peternel2017unifying}. To the best of our knowledge, unified speed-accuracy and cost-benefit trade-offs, as well as a higher-level movement-splitting strategy, have not yet been incorporated into robot control systems for collaborative human-robot movements. \rev{The most comparable approach in the human-robot collaboration literature are methods based on minimum jerk trajectories (e.g., \cite{dallard2023synchronized}). However, minimum jerk models do not enable some key features, such as 1) accounting for stochastic dynamics, which is important for goal inference, 2) accounting for the cost-benefit tradeoff that is key in predicting accurate trajectory distributions and velocity profiles, and 3) reproducing asymmetry in the velocity profile.}

To address this gap, we propose a novel robot control method that incorporates human motor control models of speed-accuracy and cost-benefit trade-offs as cost terms in optimization-based trajectory planning and higher-level movement-splitting strategy. By formulating a numerical optimal control problem that describes human reaching behaviors, this problem can be readily extended to formulate natural robot collaborative behaviors. We apply this trajectory planning to two collaborative manipulation tasks, where the goal location has higher uncertainty in certain degrees of freedom, and where the transition of the role between human and robot \rev{occurs}. The objectives (and contributions) of this study are:
\begin{enumerate}
\item[\bf  O1:] Robot motion planning that incorporates human speed-accuracy and cost-benefit trade-offs, validated by Fitts' law and velocity profiles.
\item[\bf O2:] Human-robot authority handover system that incorporates human motor control sub-movements, validated by transition point adaptation to goals of varying distance/difficulty.
\item[\bf O3:] Goal inference system which applies the same model, adding constraints of observed motion, for the online estimation of remaining trajectory, validated by approaching a variety of goals and seeing the robot adapt.
\end{enumerate}
Two applications are used for the validation of these objectives. O1 and O3 were validated in experiments involving a task that requires synchronization of human-robot co-manipulation during an object transportation task. O1 and O2 were validated in experiments involving a human-robot co-manipulation performing authority handover during an object transportation task. 

\section{Human motor control model\label{sec:human_motor_control}}
This section introduces two models for goal-oriented reaching behaviors in human motor control. \rev{ The cost-benefit and speed-accuracy trade-off model is a low-level trajectory model for human movements, and the multi-component reaching model describes when the human would switch between an initial ballistic and final corrective fine-positioning movement strategy. These two elements are also modular, where the multi-component reaching model can be employed as an authority handover between humans and robots.}

\subsection{Speed/accuracy and cost/benefit trade-offs}
\label{sec:speed_acc_tradeoff}
Human motor control optimizes trade-offs between speed and accuracy \cite{tanaka2006optimization,guigon2008computational} as well as cost and benefit \cite{rigoux2012model,berret2016don,shadmehr2016representation}, which can be explained as maximizing the following objective
\begin{align}
    J\left(\tau(t)\right) & = \mathbb{E}_{q,\dot{q},\tau} \int_0^\infty \bigg( e^{-t/\gamma} R\left(q,g\right) - \nu \Vert \tau(t) \Vert^2 \bigg) dt, \label{eq:human_obj} \\
    R(q,g) & = \begin{cases} 1 \,\,\,
    \mathrm{if}\, |x_{ee}(q)-g| < W \\ 0 \,\,\, \mathrm{otherwise}, \end{cases} \label{eq:human_reward}
\end{align}
where $q\in\mathbb{R}^{n_q}$ is human joint position and $\dot{q}$ velocity, $x_{ee}(q)\in\mathbb{R}^3$ the end-effector position, $g\in\mathbb{R}^3$ is the goal position, $t$ is the time of the movement with $t=0$ at the beginning of motion. $W\in\mathbb{R}^3$ is the goal radius (i.e., goal size), which determines the required accuracy. $\gamma$ is a discount factor for the movement time and modulates the perceived benefit of the movement. $\nu$ is a weighting factor for the metabolic cost of movement, which is determined by joint torques $\tau$. The expectation is taken with respect to the noise in the motor control system, which enters in $\tau$ and induces a distribution in $q,\dot{q}$.

This objective is subject to dynamics
\begin{align}
    M(q)\ddot{q} + D(\dot{q}) + G(q) = \tau(I+\epsilon) \label{eq:human_dyn},
\end{align}
where $\tau$ is the joint torques, while $M,\, D\in\mathbb{R}^{n_q}$ are arm inertia and damping matrices and $G(q)$ the gravitational terms. Noise $\epsilon\sim\mathcal{N}\left(0, \kappa \right)$ is the neural noise that amplifies proportionally with the muscle activity with diagonal covariance $\kappa\in\mathbb{R}^{n_q \times n_q}$.

\begin{figure}[!t]
    \includegraphics[width=\columnwidth]{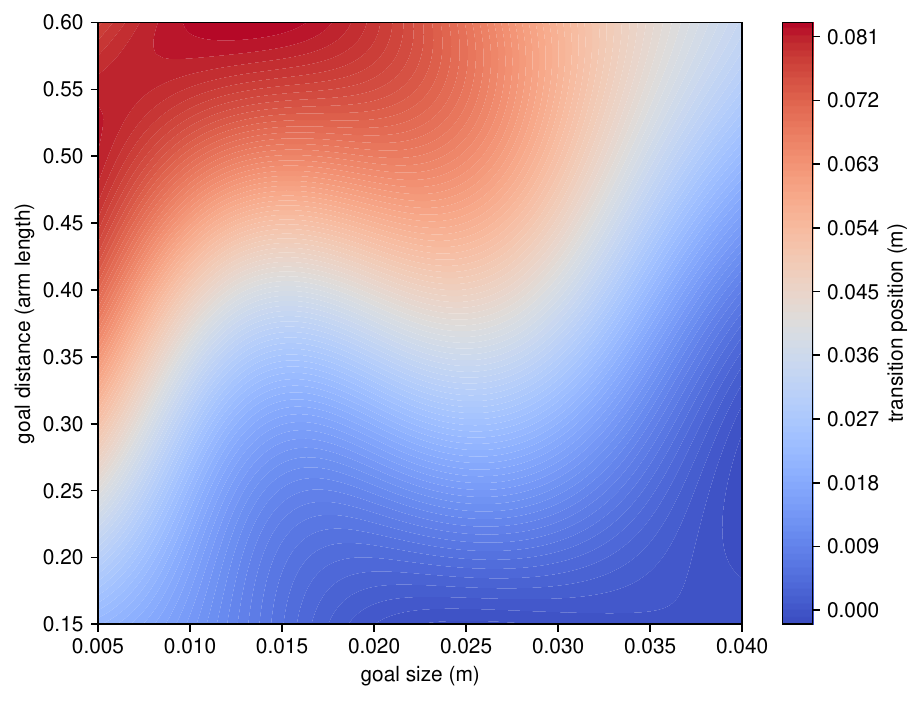}
    \caption{Multi-component strategy model for transition point prediction using Gaussian Process Regression trained on the experimental data from~\cite{peternel2019target}. The transition position expressed as is the distance from the goal and depends on the goal size and distance. The goal distance from the initial point is normalized to the arm length.}
    \label{fig:transition}
\end{figure}

\subsection{Multi-component reaching movements}
\label{sec:multi_component}
Reaching motions can be considered in two phases; an initial (ballistic) sub-movement which is faster and less precise, and a final (corrective) sub-movement is slower and more precise~\cite{elliott2001century,peternel2019target}. While the initial sub-movement is affected by visual feedback and is not entirely open-loop~\cite{elliott2010goal}, the muscular forces and timing are strongly determined by advanced task information and an internal model. These initial reaching motions are refined over trials for a specific task~\cite{elliott2004learning}.  The final sub-movement, corrective or limb-goal control, integrates visual feedback to reduce the observed positioning error between the end-effector and the goal. \rev{We consider the trajectory in both phases as modelled with the speed/accuracy and cost/benefit tradeoffs from \ref{sec:speed_acc_tradeoff} with different parameters. }

The transitions between sub-movements are affected by a range of factors including age and experience~\cite{elliott2004learning}, but are also affected by task characteristics~\cite{peternel2019target}. The transition point is assumed to be the point of highest dispersion along a point-to-point reaching task, \st{and the point of transition in relationship with} \rev{which depends on} distance and difficulty (e.g. goal width)~\cite{peternel2019target}. We used the empirical data to generate a transition point model using Gaussian Process regression (see Fig. \ref{fig:transition}). \st{ For co-manipulation HRI applications, we assume that the human presents a higher sensitivity to positioning errors in the final corrective sub-movement, and thus propose that transitions in robot behavior can occur near this transition point. }
\rev{
While these studies were performed on humans moving without collaboration, we believe that the insights are still beneficial. When applying this in human-robot co-manipulation tasks where the robot handles higher-speed motion and the human handles the adaptation~\cite{fitts1951human}, the human authority should ideally begin when non-collaborative human motor control would switch from a ballistic to a corrective strategy. We argue that this is where humans would naturally expect the strategy to change, and the human may be more sensitive to positioning errors in the corrective phase.
}

\section{Human motion from numerical optimization}
This section adapts the human motor control models from Sec. \ref{sec:human_motor_control} to make them tractable for discrete-time numerical optimization problems.

\subsection{Discretized human dynamics}
The dynamics \eqref{eq:human_dyn} are in continuous time, for a tractable numerical optimization problem we discretize the dynamics. For closed-form solutions to the expectation in \eqref{eq:human_obj}, we consider linear Gaussian dynamics. 

Denote state $s=[q^T,\dot{q}^T]^T$, $s\in\mathbb{R}^{2n_q}$, and assume the state at timestep $n$ is distributed as $s_n\sim\mathcal{N}(\mu_n,\Sigma_n)$. Then, \eqref{eq:human_dyn} can be linearized and integrated with forward Euler to find 
\begin{align}
\begin{split}
\mu_{+} & = A \mu + B \tau \\
\Sigma_{+} & = A \Sigma A^T + B \tau\left(1+\kappa\right) \tau^TB^T \\
A & = \begin{bmatrix} I & hI \\ 0 & I-hM^{-1}D \end{bmatrix} \\
B & = \begin{bmatrix} 0 \\ hM^{-1} \end{bmatrix},
\end{split}
\label{eq:numerical_dyn}
\end{align}
where $h$ is the length of the time step in seconds, i.e. $t=hn$, and the matrix $M$ is time-varying with timestep subscript suppressed. 

\subsection{Transcription of reward function}
For the optimization objective, a reward function with continuous derivatives is needed, which requires modifying the typical binary indicator \eqref{eq:human_reward}, which is $1$ within width $W$ of the goal $g$ and $0$ elsewhere~\cite{peternel2017unifying}. The reward function should also still be parameterized by width to capture task difficulty. Ideally, the reward should also allow a closed-form expectation over the Gaussian human state. 

We consider a stochastic goal ${g}\sim \mathcal{N}(\overline{g}, W)$, and take the reward as the probability that the end-effector position $x_{ee}$ matches the goal $g$, i.e. $P(x_{ee}(q)={ g})$ as
\begin{eqnarray}
R(x_{ee}) & = \frac{1}{\sqrt{2\pi|W|}}\exp\left(-\frac{1}{2}\Vert x_{ee}-\overline{g} \Vert^2_{W^{\mathtt{-}1}}\right), \label{eq:reward_conditional}
\end{eqnarray}
where $\Vert x \Vert_M^2 = x^TMx$.

In the planning problem, the human state is not exactly known; it is subject to noise which induces a distribution over $s\sim\mathcal{N}(\mu, \Sigma)$. Taking the expectation of \eqref{eq:reward_conditional}, we can find \eqref{eq:reward_conditional} as a function of $\mu, \Sigma$. We first translate the joint state distribution to a distribution over $x_{ee}$. Let the kinematics Jacobian $J = \partial x_{ee} / \partial q \in \mathbb{R}^{3 \times n_q}$, the linearized distribution of $x_{ee} \sim  \mathcal{N}(x_{ee}(\mu_q), J \Sigma_q J^T)$, where $\mu_q$ and $\Sigma_q$ are the first $n_q$ entries in $\mu$ and top-left $(n_q\times n_q)$ block of $\Sigma$. 

Denoting $\mu_x = x_{ee}(\mu_q)$ and $\Sigma_x = J \Sigma_q J^T$, the expectation of the reward function can now be found in closed form as
\begin{align}
\begin{split}
R(\mu, \Sigma) = & \mathbb{E}_{x\sim \mathcal{N}(\mu_x, \Sigma_x)}\left[ R(x) \right] \\
%= & \int_{-\infty}^{\infty} R(x_{ee})p(x_{ee}) dx_{ee} \\
= & \frac{1}{\sqrt{2\pi\left|\Sigma_x+W\right|}}\exp\left(-\frac{1}{2}\Vert  \mu_x - \overline{g} \Vert_{(\Sigma_x+W)^{-1}} \right) \label{eq:numerical_reward}
\end{split}
\end{align}
where a detailed derivation is given in Appendix~\ref{appendix}. 

\subsection{Planning problem}
With the discretized dynamics and adjusted reward function, the human motor control planning objective over a planning horizon of $H$ as
\begin{align}
    J(\tau_{0:H}) = & \sum_{i=0}^{H} \bigg( \gamma^{i-t} R(\mu_i, \Sigma_i) + \nu \Vert \tau_i \Vert^2 \bigg).
\end{align}

The planned trajectory from a state $s_n$ is then written as 
\begin{align}
\tau_{0:H} =  & \arg\min J(\tau_{0:H}) \label{eq:mpc_obj}\\
\mathrm{subject\,to}\,\,\, & \mu_0 = s_0,\,\,\, \Sigma_0 = 0,\,\,\, \eqref{eq:numerical_dyn} \label{eq:mpc_ic}
\end{align}
where \eqref{eq:mpc_ic} imposes the initial conditions and dynamics \eqref{eq:numerical_dyn}.

\begin{figure}[!t]
\centering\includegraphics[width=0.85\columnwidth]{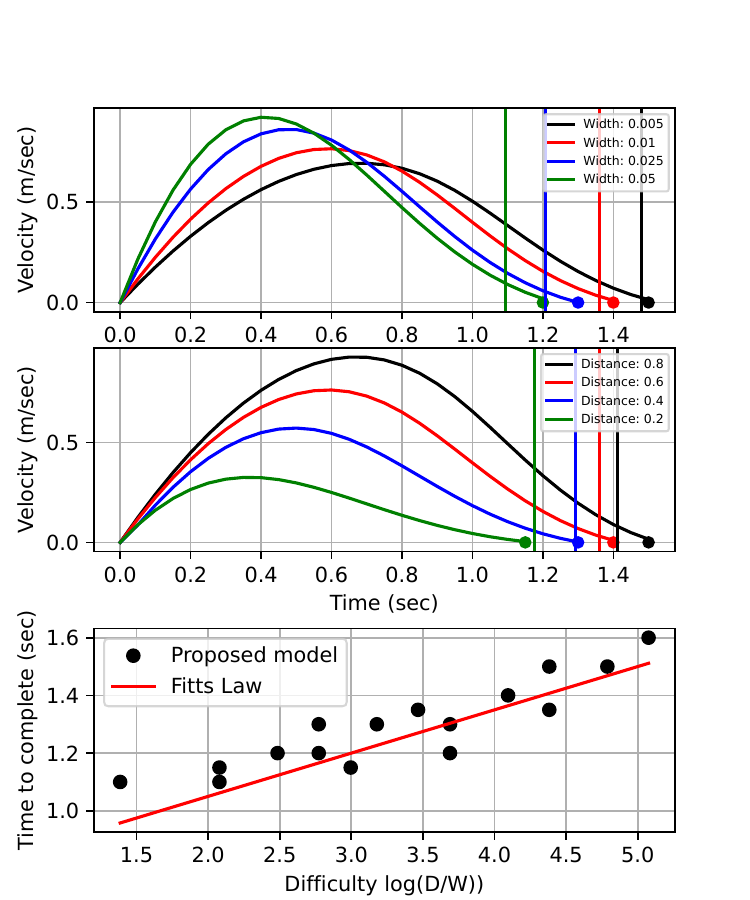}
\caption{\rev{The top two graphs show the velocity profiles of reaching movements at different target widths (which changes the task difficulty) and distances resulting from the proposed approach.  The matching vertical lines show the predicted finish time from Fitts' law. A direct comparison of finishing times} from the proposed model and Fitts' law is shown on the bottom \rev{using parameters as described in Section \ref{sec:validation}.} \st{The reaching movements were generated for varying conditions in terms of goal width and goal distance. Fitts' law graph was generated from the reaching movement data.}}
\label{fig:sim_vel_compare}
\end{figure}

\subsection{Validation of Computational Model}
\label{sec:validation}
We now verify if the proposed model adheres to both the speed-accuracy trade-off and the cost-benefit trade-off. To do so, we follow the analysis steps from the human motor control study in~\cite{peternel2017unifying} by examining velocity profiles in varying goal widths and distances, Fitts' law, and dispersion of hits at the goal over multiple reaching attempts. We validate this model on a point-to-point reaching task with a planar linear model where $M=2I$, $\overline{g}=[0.3, 0]$, and parameters seen in the `planar sim' row of Tab. \ref{tab:model_params}. The planning problem is solved at time $n=0$, and executed open-loop, with simulated muscle noise added.

The top two graphs of Fig. \ref{fig:sim_vel_compare} show the effect of goal width on the velocity profile prediction from the proposed model. We can see that the profiles exhibit key characteristics of the speed-accuracy trade-off, where velocities change based on the goal width and distance~\cite{mackenzie1987three,plamondon1997speed,peternel2017unifying,shadmehr2019movement}. For larger goals, the difficulty is lower and movements can be faster (i.e., lower movement time) and less precise (i.e., more neural noise can be accumulated throughout the movement). For smaller goals, the difficulty of hitting them is higher, and thus the movements have to be slower (i.e., higher movement time) to make sure they are precise enough in the presence of neural noise. A similar trend can be observed for goal distance, where a shorter distance represents lower difficulty and movement times are lower, while a longer distance is more difficult and movement times are higher. These trends are further illustrated in the bottom graph of Fig. \ref{fig:sim_vel_compare} where the data is calculated in terms of Fitts' law, which states that movement time increases with increased difficulty~\cite{fitts1954information}.

\begin{figure}[!t]
    \includegraphics[width=\columnwidth]{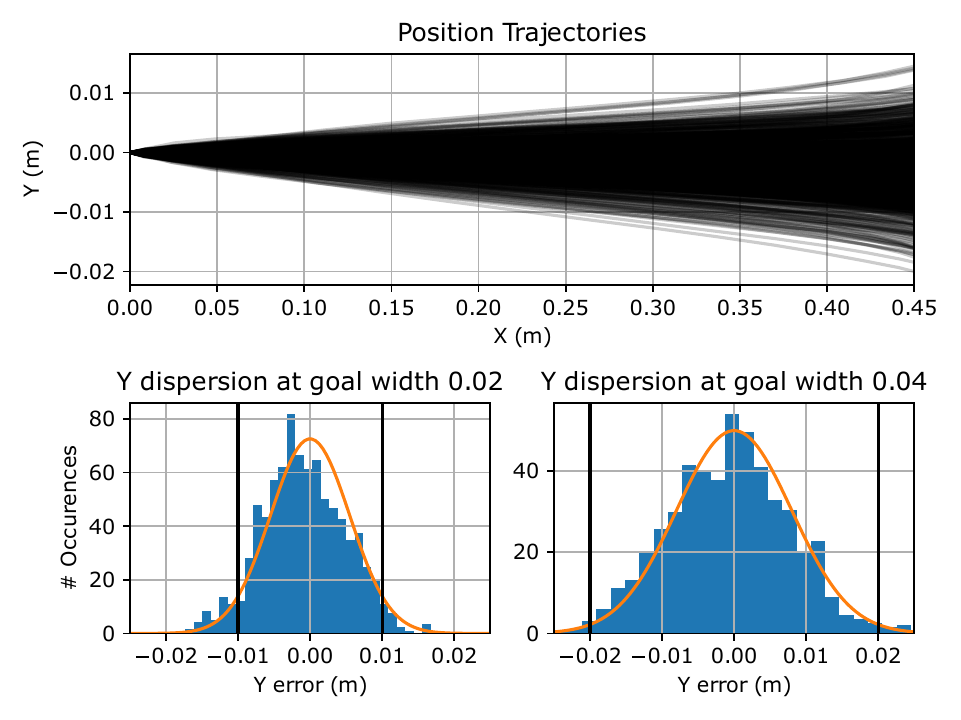}
    \caption{Dispersion of trajectories after repetitive reaching from initial position at $[0, 0]$m to the goal at $[0.45, 0]$m (top). Distribution of final $y$ values at the goals with two different widths (bottom), where the blue bars represent the number of hits at different locations, while the orange curves are fitted Gaussians with covariance $2.8e^{-2}$ and $5.4e^{-2}$ for goal width 0.02m and 0.04m, respectively. \rev{The dispersion develops along the trajectory, and final dispersion distribution matches well to results in human motor control.}}\label{fig:sim_positioning}
\end{figure}

One of the main characteristics of the cost-benefit trade-off is visible with hit dispersion on the goal in Fig. \ref{fig:sim_positioning}. The model permits a few misses in the repeated execution of movements to maximize the perceived benefit in terms of time invested into it. If the whole hit distribution were to strictly fit within the goal boundaries, the movements would have to be extremely slow so that accumulated noise would be low enough. By permitting a few misses while still keeping most in the goal, individual movements take less time and the task can be executed more often to cumulatively achieve more reward for the energy spent ~\cite{peternel2017unifying}. Additionally, Fig. \ref{fig:sim_vel_compare} can also be used to confirm some characteristics of the cost-benefit trade-off. Since investing more time into movement discounts its perceived benefit, the movement speed is adapted to the required difficulty. That is, the movement should be fast enough to minimize time spent, but not too fast, to not spend excessive energy.

Another important aspect that can be confirmed from velocity profiles in Fig. \ref{fig:sim_vel_compare} is that the model produces a temporal asymmetry in the velocity peak, which appears earlier in the movement. This temporal asymmetry is also observed in human movements~\cite{mackenzie1987three,peternel2017unifying} and can be attributed to the strategy of splitting movement into sub-movements where more time has to be invested in the later corrective stage~\cite{hansen2009three,peternel2019target}. \rev{Accounting for this asymmetry is crucial in synchronizing human-robot movements to prevent leads or lags that would hinder collaboration (e.g., unnecessary interaction forces, waiting for the robot).
}
\section{Trajectory Planning Problem Statements}
A major advantage of the formulation of dynamics \eqref{eq:numerical_dyn} and objective \eqref{eq:numerical_reward} is the ability to directly apply it to optimization-based robot trajectory planning problems. This section presents linear, nonlinear and constrained optimization-based trajectory planning problems.

\subsection{Start and end conditions}
\begin{figure}[!t]
    \centering
\includegraphics[width=0.85\columnwidth]{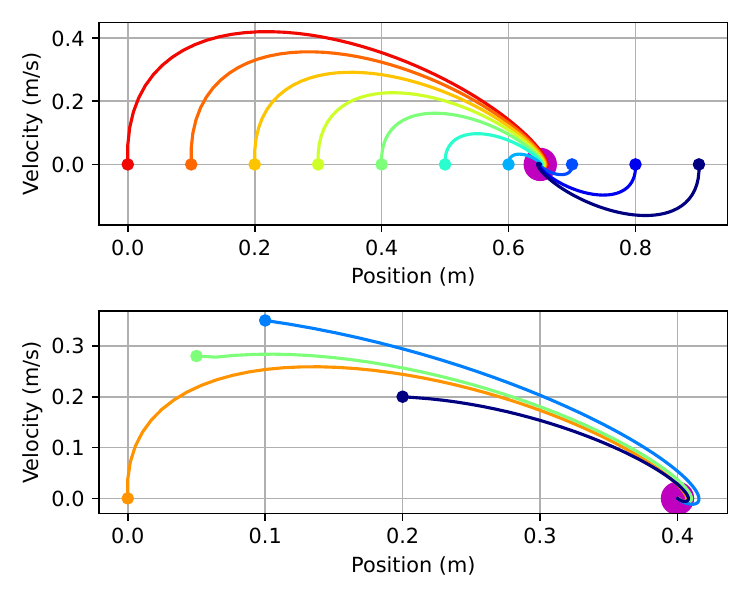}
    \caption{Position/velocity phase plots from different goal distances (top) and different initial velocities (bottom). It can be seen that the planning problem results in reasonably scaled trajectories over a range of start conditions, and is capable of handling non-zero initial velocities.}
    \label{fig:sim_phase_plot}
\end{figure}

The model should be capable of handling a range of start and end conditions, including a range of distances to goal and non-zero starting velocities. To evaluate this robustness we tested the developed robot motion planner on a variety of initial positions/velocities. The results are shown in Fig. \ref{fig:sim_phase_plot}, where the position and velocity in the $x$ direction are plotted as a phase plot. It can be seen that the proposed planner is able to adapt to different goal distances and initial velocities. Velocity is scaled as the goal distance changes according to the speed-accuracy and cost-benefit trade-offs. Furthermore, a non-zero starting velocity results in similar velocity profiles to the zero velocity starting condition.

\begin{table*}[!h]
\centering
\caption{Model and cost parameters, where $D=0.3I$, $\sigma_\tau=1e2$, unless otherwise noted.}
\begin{tabular}{r|c|c|c|c|c|c|c}
Condition & DOF ($n_q$) &  Width [m] & Discount $\gamma$ & Action cost $\nu$ & Steps $H$ & Step time $h$ &  Solve time (s)  \\
\hline
Planar sim. & $2$ &$[0.005, ... 0.04]$ & $[0.97,... 1.0]$ & $1e-5$ & $30$ & $0.02$ & $0.04$ \\
Cartesian & $3$ &$0.05$ & $0.97$ & $1e-5$ & $50$ & $0.055$ & $0.20$ \\
Franka Emika & $7$ & $0.05$ & $0.97$ & $1e-5$ & $50$ & $0.035$ & $1.3$ \\
\end{tabular}
\label{tab:model_params}
\end{table*}

\subsection{Scaling to nonlinear dynamics}
We also apply the problem to nonlinear system dynamics, using the inertial model of the Franka Emika robotic arm. This presents a nonlinear $M(q)$ and $J(q)$. The optimization problem can scale to this more complex model, albeit the time required to solve increases to over $1.3$ seconds, as seen in Table \ref{tab:model_params}. The resulting trajectories can be seen in the attached video.

\subsection{Goal estimation from constrained optimization}
One benefit of the optimization-based approach is flexible adaptation of the problem statement. We show this by using \eqref{eq:mpc_obj} for goal estimation. We measure a position and velocity during the motion and use this to infer the remainder of the trajectory. This is here formulated by adding the observed state $s_o$ at time $t_o$ as a constraint to the optimization problem, then optimizing the trajectory and goal such that the start and observed states are fixed. This extends \eqref{eq:mpc_obj} as
\begin{align}
    \tau_{1:H}, \hat{g} &= {\arg \min}_{\tau, g} J(\tau_{1:H}, \hat{g}) + \Vert \hat{g}-\overline{g}\Vert_{W^{-1}} \label{eq:goal_est}\\ 
    \mathrm{subject\,\,to}\,\,\,& s_{n_o} = s^o, \,\,\eqref{eq:mpc_ic}
\end{align}
where $s^o$ is the observed state at the discrete time-step closest to the continuous time of measurement $n_o=\lfloor t_o/h \rfloor$, and $\hat{g}$ is the estimated goal.  Note that a regularization on the estimated goal has been added.

\section{Collaborative application}
\begin{figure}[!t]
    \includegraphics[width=\columnwidth]{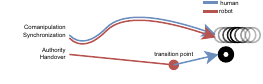}
    \caption{Two human-robot co-manipulation scenarios in which we showcase the application of the developed robot online motion planner based on the human motor control models. The top scenario shows a case where different degrees of freedom controlled separately by either the human or the robot must be synchronized in the face of an uncertain goal. The bottom scenario shows a case where the robot leads the execution of one stage of the task and the human another stage, thus the key is to infer the transition of authority.}
    \label{fig:application_concept}
\end{figure}

\begin{figure}[!t]
    \centering
    % \subfloat[Synchronization task]{\includegraphics[width=0.99\columnwidth]{figs/exp_setup_cable.JPG}}\\
    % \subfloat[Authority transition task]{\includegraphics[width=0.99\columnwidth]{figs/exp_setup_hob.JPG}}
    \includegraphics[width=\columnwidth]{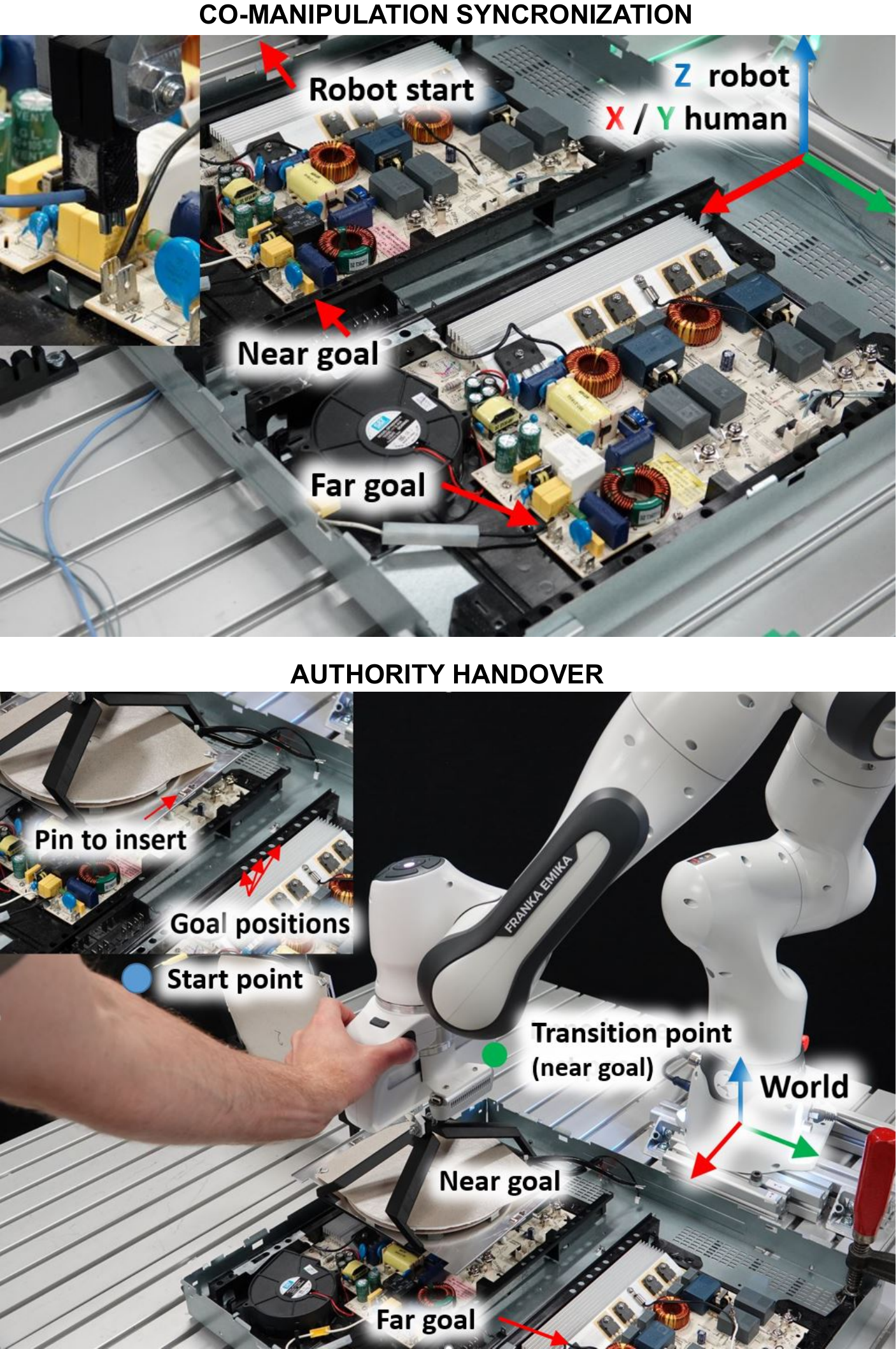}
    \caption{Experimental validation on an electronics assembly use case using a collaborative Franka Emika robotic arm. The top image shows the co-manipulation synchronization scenario where there is higher task variation in the $x-y$ plane and a more deterministic movement in the $z$-axis\rev{, therefore these are human- and robot-lead, respectively}. The bottom image shows the authority handover scenario where fine positioning is required after the transition point\rev{, with the robot start position and the two goal positions used in Section \ref{sec:handover} shown}.}
    \label{fig:exp_setup}
\end{figure}

We investigate the feasibility of the human motor control principles in collaborative strategies by applying them to two collaborative tasks within an electronics assembly use case. The two collaborative tasks are schematically presented in Fig. \ref{fig:application_concept}, while the experimental setup and scenarios are shown in Fig. \ref{fig:exp_setup}. In both scenarios, the robot and human must physically collaborate to accomplish an assembly task which includes variation in the final position, requiring human input to succeed. A Franka Emika robotic arm is used with impedance control, where \rev{the impedance rest position follows the proposed approach,} the rotational stiffness is kept high, \st{while reference position }and Cartesian stiffness\st{are varied according to the proposed approaches} \rev{is switched from high to $0$ according to the proposed approaches}. 

\subsection{Co-manipulation synchronization}
\label{sec:sync}
In this scenario, we consider an assembly task where certain degrees of freedom involve task variation and therefore require adaptability, while the complementary degrees of freedom are deterministic and can benefit from robotic precision. An example of such a task is collaborative transportation, where the height trajectory is known, while the trajectory in the horizontal plane should be adapted. Due to superior cognitive capabilities and adaptability, humans should be assigned to control the degrees of freedom with task variation. On the other hand, the deterministic ones can be controlled by the robot, which has superior precision and load capacity. The challenge of this problem is how to make a robot synchronize its motion to the human motion to match the peak velocity and time to complete\rev{, to reduce motion conflicts between robot and human}. 

A collaborative assembly task can be seen in Fig. \ref{fig:exp_setup}(a), where the height of the table is fixed and known, but the \rev{goal} location in $x-y$ plane\rev{, on the table,} varies. The robot starts with zero translational stiffness \rev{at an arbitrary start position}, and the human initiates the co-manipulation \rev{by grasping and moving towards the current goal}.  At $t_o=0.2$ seconds after motion is started, the position and velocity are measured to give the observed state $s_o$. The problem \eqref{eq:goal_est} is solved \rev{with the initial state and observed state $s_0$.} The resulting Cartesian trajectory is sent as the rest position of the impedance controller, and the translational stiffness is set to $[0,0,400]$N/m. \rev{The trajectory is executed without re-planning, and the operator can freely manipulate the $x$ and $y$ directions, and is driven along the robot's trajectory in $z$ with $400$ N/m stiffness.}

\begin{figure}[!t]
    \centering\includegraphics[width=0.85\columnwidth]{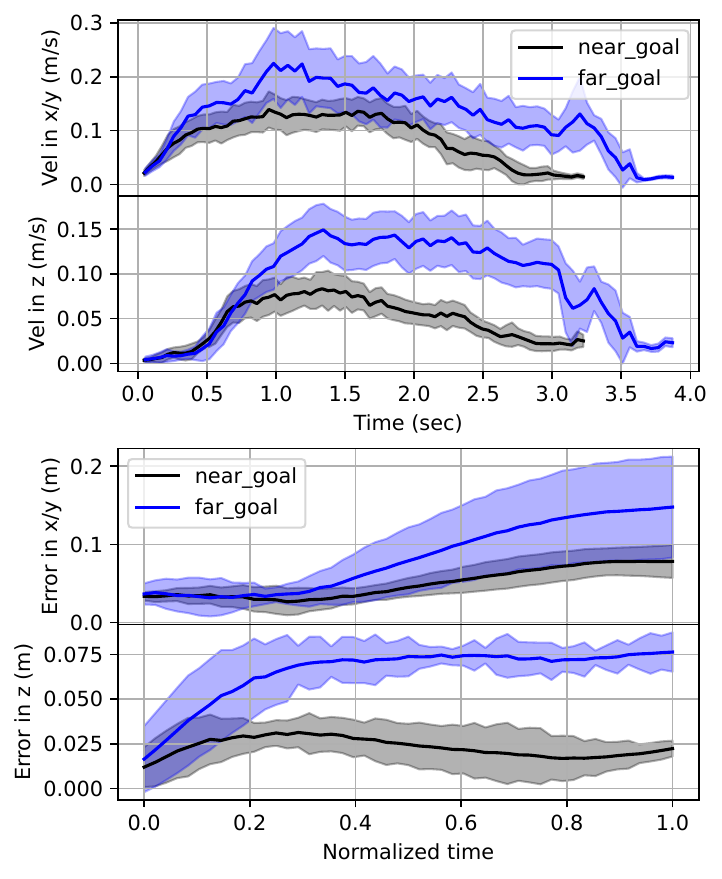}
    \caption{Resulting velocity profiles over time (top) of the synchronized motion, and (bottom) error between estimated Cartesian trajectory and actual robot end-effector position over normalized time. \rev{Mean and standard deviation are shown over 5 trials with a single operator.}}
    \label{fig:sync_results}
\end{figure}

\rev{An operator with robotics experience but otherwise not involved in the project approaches two goals, with 5 iterations each.} Figure \ref{fig:sync_results} shows the results of synchronized human-robot co-manipulation for two goals at different distances\rev{, with mean and standard deviation over the trials}. The velocity profile in Fig. \ref{fig:sync_results}(a) is smooth even when the robot estimates and assists, and is largely similar to human trajectory profiles without active robot assistance. We can see that both human and robot velocity profiles exhibit temporally asymmetric velocity peaks, where the peak comes closer to the beginning of the movement and changes based on the goal distance. Furthermore, the times when the goal is reached are matched by different degrees of freedom.

The error between estimated human trajectory and real robot position over time in Fig. \ref{fig:sync_results}(b) is larger in the $x-y$ plane than in the $z$ direction, and is larger when the goal is farther. This is in line with the speed-accuracy trade-off. \rev{Additional $z$ error is caused from the object weight $1.5$kg, which is not compensated.} The ability to handle goals in novel directions is shown in the video attached to the supplementary material or at this url: \url{https://youtu.be/9caL9bBedA0}.

\subsection{Authority Handover}
\label{sec:handover}
In this scenario, we investigate a case when human input may be only necessary for fine-positioning (e.g., corrective motion during the precise assembly). An example of such a task is collaborative assembly where the part has to be brought into the vicinity of the assembly point quickly, after which task variation and complexity require slower adaptive motion. In this case, the robot can handle the initial transportation motion before transferring the authority for final positioning to the human. The main challenge is for the robot to know when and where to hand over the authority. To solve this challenge, we propose using insights from the human multi-component movement strategy to determine when the human would typically transition from the initial to the final movement component.

In our approach, in the beginning, the robot produces a trajectory based on human-inspired initial sub-movement, i.e., using the planner in \eqref{eq:mpc_obj} and \eqref{eq:mpc_ic}, and parameters `Cartesian' in Table \ref{tab:model_params}. This trajectory is sent as rest position to an impedance controller with translational stiffness of $400$N/m. The robot then uses a human multi-component strategy model to infer the timing and location of the transition point between the initial to final sub-movement to transition the authority to the human where the fine-positioning authority typically occurs. When the transition point is reached, the robot translational stiffness is set to $0$ in order to allow the human to take over leading the final motion. To infer that transition point, we use the model from Sec. \ref{sec:multi_component}, seen in Fig. \ref{fig:transition}.

\begin{figure}[!t]
    \includegraphics[width=0.85\columnwidth]{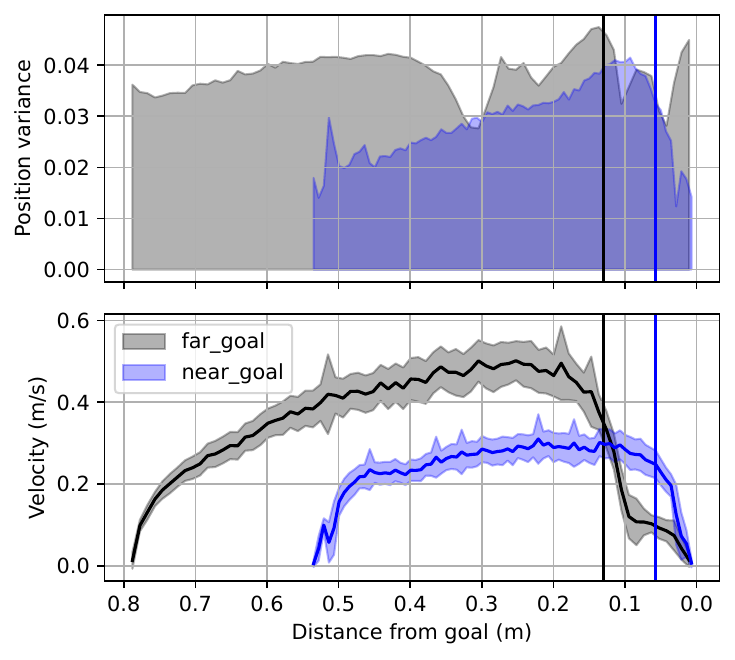}
    \caption{Position variance vs distance to goal (top), \rev{velocity mean and standard deviation vs. distance to goal} for the various goal distances. The predicted sub-movement transition point is marked with a vertical line, \rev{results are shown for 5 trials with a single operator}.}
    \label{fig:switch_comp}
\end{figure}

\rev{The proposed approach is applied with a test subject who has robotics experience but is otherwise not involved with this development, alternating between near and far goal. Each goal is approached 5 times.} Figure \ref{fig:switch_comp} shows this transition behavior for a goal which is near and far from the robot starting position. The \rev{standard deviation} in position and \rev{mean and standard deviation of the} velocity profiles are plotted over the distance to the goal, while the transition points are shown in vertical lines.  On the top graph, we can see that the peak dispersion for each goal is close to the respective predicted transition point. On the bottom graph, it can also be seen that no jump in velocity occurs at the transition point, which indicates a smooth transition was achieved.  %Furthermore, the human forces are higher in the fine positioning phase since fine adjustments had to be made.

\section{Discussion}
This paper adapted two models of human motor control to collaborative robotics for the purpose of human intent estimation and robot online trajectory planning. This is the first study to integrate unified speed-accuracy/cost-benefit trade-offs and human multi-component strategies into a robot planner for online human intent estimation. The results show that adapted models match the expected properties of basic human motor control and have the solve time and generality over start/end conditions to be suited for online planning. We showcased the flexibility of the proposed approach by validating it in two scenarios: synchronization of human-robot co-manipulation and handover of authority from robot to human.

The study objective O1 was validated by computational analysis in Sec. \ref{sec:validation}. We showed that the developed robot planner accounts for all the major aspects of speed-accuracy and cost-benefit trade-offs, consistent with the human motor control literature. This objective was further validated by experimental analysis in Sec. \ref{sec:sync} where we see that velocity profiles follow similar patterns. The study objective O2 was validated by experimental analysis in Sec. \ref{sec:handover}. The results showed that the robot planner could use the human multi-component strategies model to infer the transition point for handover from robot to human. The study objective O3 was validated by experimental analysis in Sec. \ref{sec:sync}. The results showed that the developed goal inference system can predict changes in goal location and inform the planner to adjust the robot's trajectory. The velocity profiles were adapted based on the inferred goal. 

\st{The purpose of this study was to create a robot motion planner based on human motor control models and validate it experimentally on realistic human-robot collaboration tasks.} Some limitations of the approach are from the motor control models employed: they are only point-to-point motions and do not consider interaction dynamics. New models from human motor control may improve these aspects\st{and modularity of the proposed framework enables the incorporation of any future advancements in the field of human motor control}. On the robotics side, the approach still requires an initial guess for the human goal, the full nonlinear solve requires $>1$ second, and \st{the cost function has weights to} \rev{parameters like goal width and cost function weights must} be tuned. Additionally, the effect of motor control-based trajectories on subjective perceptions was not studied, thus a future step includes user studies to investigate this.

\appendices 
\section{Expected reward}
\label{appendix}
The integrand in the expectation
\begin{eqnarray*}
\mathbb{E}_{x\sim \mathcal{N}(\mu_x, \Sigma_x)}\left[ R(x) \right]  & = \int_{-\infty}^{\infty} R(x_{ee})p(x_{ee}) dx_{ee}
\end{eqnarray*}
is the product of two Gaussian densities in $x_{ee}$, $R(x_{ee})=\mathcal{N}(x_{ee};g, W)$ and $p(x_{ee}) = \mathcal{N}( x_{ee};\mu_x, \Sigma_x)$, where
$\mathcal{N}(x;\mu,\Sigma)=\vert 2\pi\Sigma\vert^{-1/2}\exp\left(-\frac{1}{2}\mu^T\Sigma^{-1}\mu\right)$.
The product can be found \st{be}\rev{by} \cite[(371)]{petersen2008} as
\begin{multline*}
    \mathcal{N}(x_{ee};g, W) \mathcal{N}( x_{ee};\mu_x, \Sigma_x) = N_c\cdot\\
      \mathcal{N}(x_{ee}; (\Sigma_x^{\mathtt{-}1}+W^{\mathtt{-}1})^{\mathtt{-}1}(\Sigma_x^{\mathtt{-}1}\mu_x + W^{\mathtt{-}1}g), (\Sigma_x^{\mathtt{-}1}+W^{\mathtt{-}1})^{-1})
\end{multline*}
where~$N_c=\vert 2\pi(\Sigma_x + W) \vert^{-1/2} \exp \left( -\frac{1}{2}
\Vert \mu_x-g \Vert_{(\Sigma_x+W)^{-1}} \right)$.
Recalling that the integral over the multivariate Gaussian is $1$, we find
\begin{eqnarray*}
    R(\mu, \Sigma) = \frac{\exp \left( -\frac{1}{2}
\Vert \mu_x-g \Vert_{(\Sigma_x+W)^{-1}} \right)}{\sqrt{\vert 2\pi(\Sigma_x + W) \vert}}.
\end{eqnarray*}

\bibliographystyle{IEEEtran}
\bibliography{lib,extrarefs}

\end{document}